\documentclass[10pt,twocolumn,letterpaper]{article}

\usepackage{iccv}
\usepackage{times}
\usepackage{epsfig}
\usepackage{graphicx}
\usepackage{amsmath}
\usepackage{amssymb}
\usepackage{multirow}
\usepackage{orcidlink}
\usepackage{subcaption}
\usepackage{hyperref}
\usepackage{cleveref}
\iccvfinalcopy % *** Uncomment this line for the final submission

\def\iccvPaperID{20} % *** Enter the ICCV Paper ID here

\ificcvfinal\fi

\begin{document}

%%%%%%%%% TITLE
\title{CNN based Cuneiform Sign Detection Learned from Annotated 3D Renderings and Mapped Photographs with Illumination Augmentation}

\author{Ernst Stötzner$^{1}$ \orcidlink{0009-0003-6716-2979}, Timo Homburg$^{2}$ \orcidlink{0000-0002-9499-5840}, Hubert Mara$^{1}$ \orcidlink{0000-0002-2004-4153}\\
\and
$^{1}$Institute of Computer Science\\
Martin-Luther-Universität Halle-Wittenberg, Germany\\
{\tt\small https://www.informatik.uni-halle.de/}
\and
$^{2}$i3mainz – Institute for Spatial Information \& Surveying Technology\\
University of Applied Sciences Mainz, Germany\\
{\tt\small https://i3mainz.hs-mainz.de/}
}
%\author{Ernst Stötzner \orcidlink{0009-0003-6716-2979}\\
%Institute of Computer Science\\
%Martin-Luther-Universität Halle-Wittenberg, Germany\\
%{\tt\small https://www.informatik.uni-halle.de/}
%\and
%Timo Homburg \orcidlink{0000-0002-9499-5840}\\
%i3mainz – Institute for Spatial Information \& Surveying Technology\\
%University of Applied Sciences Mainz, Germany\\
%{\tt\small https://i3mainz.hs-mainz.de/}
%\and
%Hubert Mara \orcidlink{0000-0002-2004-4153}\\
%Institute of Computer Science\\
%Martin-Luther-Universität Halle-Wittenberg, Germany\\
%{\tt\small https://www.informatik.uni-halle.de/}
%}

\maketitle
\ificcvfinal\thispagestyle{empty}\fi

\begin{abstract}
   Motivated by the challenges of the Digital Ancient Near Eastern Studies (DANES) community, we develop digital tools for processing cuneiform script being a 3D script imprinted into clay tablets used for more than three millennia and at least eight major languages. It consists of thousands of characters that have changed over time and space. Photographs are the most common representations usable for machine learning, while ink drawings are prone to interpretation. Best suited 3D datasets that are becoming available. We created and used the HeiCuBeDa and MaiCuBeDa datasets, which consist of around 500 annotated tablets.
   %The largest open-access benchmark dataset contains about 2,000 tablets, which have been extended in previous work with uniform renderings, metadata, and 3D meshes. About a third of the tablets have been manually annotated with bounding polygons for benchmark purposes. 
   For our novel OCR-like approach to mixed image data, we provide an additional mapping tool for transferring annotations between 3D renderings and photographs. Our sign localization uses a RepPoints detector to predict the locations of characters as bounding boxes.
   %For the localization of the characters in the form of bounding boxes, we use a RepPoints detector.
   %The OCR pipeline uses a RepPoints detector to predict the location of characters as bounding boxes. 
   %It uses a RepPoints detector with a ResNet18 as a backbone, where the resulting positions are used to compute a bounding box for detected characters using two non-shared subnetworks on the feature map input. Hyperparameter optimization is performed for the parameters of the Stochastic Gradient Descent (SGD), and Non-Maximum Suppression (NMS) is used as post-processing. 
   We use image data from GigaMesh's MSII (curvature) based rendering, Phong-shaded 3D models, and photographs as well as illumination augmentation. The results show that using rendered 3D images for sign detection performs better than other work on photographs. In addition, our approach gives reasonably good results for photographs only, while it is best used for mixed datasets. More importantly, the Phong renderings, and especially the MSII renderings, improve the results on photographs, which is the largest dataset on a global scale.
\end{abstract}
\section{Introduction}
The cuneiform script is one of the oldest writing systems in the world. The majority of the cuneiform script is found on clay tablets into which each wedge-shaped cuneiform sign was impressed with a reed stylus. Due to the three-dimensional nature of the signs, the script is only legible with proper illumination. Consequently, the work with single photographs is limited because the lighting is fixed, while reading requires repositioning the often curved tablets. So more recent approaches are based on imaging systems capturing information about the 3D shape such as the \emph{Leuven Dome}~\cite{23LeuvenDom}. Especially, \emph{Structured Light Scanning}~(SLS) is becoming increasingly popular for documentation of small archaeological findings~\cite{KarlHouskaLengauerHaringTrinklPreiner}. So 3D models captured by 3D scanners are also increasingly used to capture and visualize clay tablets with cuneiform script, especially in combination with high-quality curvature rendering technique for 3D datasets using \emph{Multi-Scale Integral Invariant}~(MSII) filtering~\cite{10MaraGigaMesh}. Filtering and rendering is applied with the Free and Open Source Software \emph{GigaMesh}\footnote{https://gigamesh.eu}.
The first steps in using neural networks to recognize cuneiform writing on images were taken in the 1990s~\cite{97Roshop}.
Applying artificial intelligence systems directly to 3D models has proven challenging, but \cite{bogacz2020} has shown promising results for period classification of tablets. 
This article, which is a contribution to \emph{Digital Assyriology}~\cite{22Bartosz} also known as \emph{Digital Ancient Near Eastern Studies}~(DANES), focuses on the use of different types of renderings and explores their potential in machine learning as a step towards \emph{Object Character Recognition}~(OCR) of this particular ancient script.
%being used for more than three millennia. 
In addition, we will compare the results on differently rendered 3D datasets in combination and comparison with photographs for the task of sign detection.

\subsection*{Related Work}

The formidable challenge of OCR for cuneiform is an important DANES research topic and paves the way for the vision of automatic ancient language translation as the recently introduced \emph{Neural Machine Translation}~(NMT) model that allows the translation of Akkadian into English by \cite{23Gutherz}. The promising results, evaluated by the test data and experts, are interesting pioneering work in automatic Cuneiform translation. 
However, their NMT requires transliterations in cuneiform Unicode or transliteration in Latin script and cannot be performed on photographs or 3D renderings. Consequently, the OCR preprocessing of cuneiform tablets is necessary. 
\cite{23deepScribe} describes an OCR system to determine transliterations which consist of several subtasks such as sign localization, sign classification, and sign-to-line assignment. This procedure is a complete pipeline from a photograph as input to a transliteration as output. This pipeline performs the above-mentioned steps of OCR separately, first locating and cropping, then classifying the extracted signs, and finally arranging them into lines. The end-to-end evaluation of the pipeline with \emph{Character Error Rate}~(CER) of $0.69$ is improvable and insufficient yet. However, the sign detection task reached an \emph{Average Precision}~(AP) of $0.78$. It is worth mentioning that their available dataset of over $1300$ fully annotated tablets is a notably larger corpus compared to research presented in~\cite{Dencker20},~\cite{Hamplova22}, and~\cite{Rest22}.

Cuneiform was used as the script of several languages, including Elamite, Sumerian, and Assyrian.  In the example of the Elamite language, an annotated dataset\footnote{DeepScribe:\url{https://github.com/edwardclem/deepscribe}} is now available. However, the availability of expert annotated datasets is still limited for other languages and periods. To contribute to the solution of the missing datasets, we provide our applied dataset of manually annotated 3D renderings under a CC-BY license. 
Because of the scarcity of data,~\cite{Dencker20} introduced a weak-supervised learning approach that uses a large dataset of transliterations and annotated photographs from the Cuneiform Digital Library Initiative (CDLI)\footnote{\url{https://cdli.mpiwg-berlin.mpg.de}} to train a cuneiform sign detector that locates and classifies the signs. Another approach reduces the impact of data limitations through illumination augmentation~\cite{Rest22}. Using a 3D model rendered under different lighting conditions to augment their dataset of cut-out signs has shown promising results in their sign classification. However, their main dataset consists of cropped signs from photographs from the Hethitologie Archiv at the Hethitologie Portal Mainz (HPM)\footnote{\url{https://www.hethport.uni-wuerzburg.de/HPM/index.php}}. 
%In contrast, our approach is to use raster images of renderings as standalone data. We also we apply a similar illumination augmentation method, but on an entire dataset of 3D models.

In this work, we use a Convolutional Neural Network (CNN) to locate cuneiform signs on uniformly sized cropped images which could be utilized in a pipeline as the recently presented in~\cite{23deepScribe}. Their results suggest a weakness in accurately predicting overlapping bounding boxes with the ground truth. As a consequence of incompletely cropping signs due to the inaccurately predicted bounding boxes, a potential information loss may occur. This could explain the decrease in classification accuracy of the whole pipeline compared to their classifier, which was evaluated on cropped signs from the annotations. Therefore, we propose to focus on evaluating the sign localization by considering true positives with an Intersection-over-Union (IoU) overlap with the ground truth bounding box of at least 75\%. We present a sign detector trained on 3D renderings that still performs well with this more strict evaluation constraint. For the evaluation, we compare different types of renderings with photographs to determine their potential for such a task. Furthermore, our evaluation considers the type of image used, the accuracy of the bounding boxes, and the effect of illumination augmentation. 
For our training, we use four available raster image datasets separately and in combination. We also apply a similar illumination augmentation method as \cite{Rest22}, but to our entire dataset of 3D models.

The aforementioned need for an accurate sign localization as part of a pipeline is further motivated by our approach, which will be introduced in the following section, along with our approach to this sign detection challenge.   
%The following section briefly introduces a pipeline and our approach to sign detection. The motivation for the need for accurate sign detection comes from this pipeline. 

\section{Method}
First, we briefly introduce a wedge detection pipeline that includes the sign detector described in~\ref{sec:methSignDet}. Second, we describe a data augmentation approach based on 3D renderings in \Cref{sec:illuminationAugmentation}.

\subsection{Wedge Detection Pipeline} \label{sec:pipeline}
\cite{our23} introduced a pipeline approach to detect wedges in images of entire Cuneiform tablets. Similar to \cite{23deepScribe}, the pipeline initially locates the signs as bounding boxes in images and crops them. These cutouts were used to detect and classify the wedges, unlike \cite{23deepScribe} where the signs are classified.
In \cite{our23}, the same architecture as in this work was used for the sign detection, but only images of whole tablet segments were used, which gave useful results with a mean $F_1-Score=0.61$ on renderings but performed less optimal for large tablets. The investigated wedge detector is based on the ideas of \emph{Point RCNN}~\cite{PointRCNN} approach. In summary, a Region
Proposal Network (RPN) based on \emph{RepPoints}~\cite{yang2019reppoints}, predicts an area in the image known as the Region of Interest (RoI) as a bounding box, which may contain a wedge, and the features of this region are extracted by RoI Align~\cite{maskRCNN}. These RoI features are the input of a refinement neural network that provides offsets for each corner point of the bounding box to specify a wedge-shaped rotated quadrilateral instead of a bounding box. Furthermore, this network classifies the wedge according to the PaleoCodage encoding~\cite{homburg2021paleocodage} or the Gottstein system~\cite{Gottstein2012EinSI}. The effect of the two systems for the wedge detection was evaluated and discussed in \cite{our23}, where the Gottstein system achieved a slightly higher mean precision of 0.52 compared to the PaleoCodage mean precision of 0.43. In general, the precision of the wedge detection appears to be sufficient to apply, \eg for an automatic alignment of the clay tablets, but the challenge is to improve on the low recall of less than 0.19 when using the PaleoCodage encoding and 0.15 for the Gottstein system.
The further developments of this important challenge are not part of this work. However, the sign detector was further investigated and improved by an augmentation method described below.

\subsection{Sign Detection} \label{sec:methSignDet}
The sign detector is a single-class object detection task, where the outputs are bounding boxes defined by $(x_{min},y_{min},x_{max},y_{max})$ with an assigned confidence value between 0 and 1, whether the bounding box is a sign or not.
As in \cite{our23}, we used the one-stage anchor-free object detector \emph{RepPoints}~\cite{yang2019reppoints} with a \emph{ResNet18}~\cite{ResNet} as backbone. 
Originally, \cite{yang2019reppoints} introduced the \emph{RepPoints} architecture with a \emph{ResNet} backbone used as a \textit{Feature Pyramid Network (FPN)} \cite{FPN}. However, we did not achieve better results with FPN, so we simplified our architecture by using the layer $c4$ of \emph{ResNet18}, which results in a feature map $f$ with a resolution of $64\times64$ pixel for our input size of $512\times512$ pixel squares.  

\emph{RepPoints} does not directly predict the bounding box, but a set of $k$ representation points and a confidence value for $c+1$ classes for each point $(x_f,y_f)$ of the feature map $f$ returned by the backbone, where $c$ is the number of classes, and one is added as the background. In our task, we set $c=1$ because we decided between sign and background.
%, but \emph{RepPoint} can also be used with more than one class. 
Applying two non-shared subnetworks with the feature map as input leads to the classifications and localizations consisting of two point sets $P_1(x_f,y_f)$ and $P_2(x_f,y_f)$ of the objects.
The set $P_1(x_f,y_f)$ is a result of $k$ offsets $\Delta x_{f}$ and $\Delta y_{f}$ for each feature map point (cf. \Cref{eq:deltaoffset}):
\begin{equation}\label{eq:deltaoffset}    
P_1(x_f,y_f) = \lbrace(x_f+\Delta x_{f_{i}},y_f+\Delta y_{f_{i}})\rbrace_{i=1}^{k}
\end{equation}
These $k$ offsets are the first part of the localization subnet and are also used as deformable convolutional layer input offsets in classification and further localization subnet. 
To refine the point positions, the localization subnet predicts the set $P_2(x_f,y_f)$ by $k$ offsets $\Delta x'_{f}$ and $\Delta y'_{f}$, based on the points in $P_1$ (cf. \Cref{eq:offsetfunc}).
\begin{equation}\label{eq:offsetfunc}
\begin{split}
P_2(x_f,y_f) =& \lbrace(x_{p_{1}i}+\Delta x'_{f_{i}},y_{p_{1}i}+\Delta y'_{f_{i}})\rbrace_{i=1}^{k},\\
&(x_{p_{1}i},y_{p_{1}i})\in P_1(x_f,y_f)
\end{split}
\end{equation}
For each experiment, $k=9$ is used, following \cite{yang2019reppoints}.
To train the network and to get a bounding box result, these $k$ points of the point set $P_j$ for each feature map position $(x_f,y_f)$ must be converted to a pseudo bounding box, where the min-max function is used for both dimensions, as shown in \Cref{eq:minmaxfunc}.
\begin{equation}\label{eq:minmaxfunc}
\begin{split}
    \hat{x}_{min}=\min_{x_p \in P_j(x_f,y_f)}(x_p) \\
    \hat{y}_{min}=\min_{y_p \in P_j(x_f,y_f)}(y_p) \\
    \hat{x}_{max}=\max_{x_p \in P_j(x_f,y_f)}(x_p) \\
    \hat{y}_{max}=\max_{y_p \in P_j(x_f,y_f)}(y_p) \\
\end{split}
\end{equation}

\begin{figure}[t]
\begin{center}
   \includegraphics[width=0.6\linewidth]{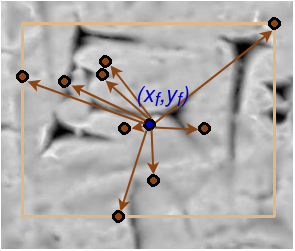}
\end{center}
   \caption{Example of the set $P_1(x_f,y_f)$(brown) with the resulting pseudo bounding box. The arrows symbolize the offset from $(x_f,y_f)$(blue) to the $k$ points. }
\label{fig:repPoints1}
\end{figure}

\begin{figure}[t]
\begin{center}
   \includegraphics[width=0.6\linewidth]{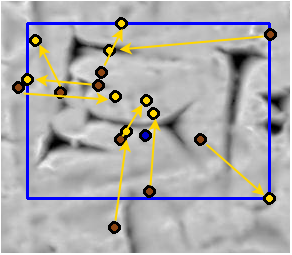}
\end{center}
   \caption{Example of the set $P_2(x_f,y_f)$ (yellow) with the resulting pseudo bounding box. The arrows symbolize the offset from $P_1(x_f,y_f)$ (brown) to the corresponding $k$ points of $P_2(x_f,y_f)$. }
\label{fig:repPoints2}
\end{figure}

The \Cref{fig:repPoints1} shows an example of representation points set $P_1(x_f,y_f)$ and the offsets $\Delta x'_{f}$ and $\Delta y'_{f}$ with the pseudo bounding box at the feature map point $(x_f,y_f)$. The refinement of these points by the second offsets $\Delta x'_{f}$ and $\Delta y'_{f}$ and the resulting bounding box, which is also the final detection result, is visualized in \cref{fig:repPoints2}. 

The training is driven by two localization losses and one classification loss. Both localization losses are calculated by the smooth $l_1$ distance between the four bounding boxes describing values $x_{min},y_{min},x_{max}$ and $y_{max}$ of the ground truth (GT) bounding boxes and the predicted pseudo bounding boxes. As described in \cite{yang2019reppoints}, the center of the GT bounding boxes are projected to the feature map position, and only for these feature map points the location loss $\mathcal{L}_{loc_{1}}$ of pseudo bounding boxes $(\hat{x}_{min},\hat{y}_{min},\hat{x}_{max},\hat{y}_{max})$ by the min-max function based on $P_1$ is calculated.
The second location loss $\mathcal{L}_{loc_{2}}$, based on $P_2$, is computed only for those feature map points where the pseudo bounding box of $P_1$ has an intersection-over-union (IoU) value with the GT bounding box above the threshold $\theta_{TP}$.
The classification loss $\mathcal{L}_{class}$, for which the Focal Loss~\cite{focalloss} is used, also depends on the bounding box of $P_1$. In addition to the threshold $\theta_{TP}$, $\theta_{FP}$ is defined, where all predicted boxes with an IoU value above $\theta_{FP}$ but below $\theta_{TP}$ belong to the GT 'background', and if the IoU value is above $\theta_{TP}$ the GT class corresponding to the bounding box is considered as 'sign'. Unlike \cite{yang2019reppoints}, we have increased the original values $\theta_{FP}=0.4$ and $\theta_{TP}=0.5$ to $\theta_{FP}=0.6$ and $\theta_{TP}=0.7$ because the signs are dense and experiments with the original values gave worse results. 
In summary, the complete loss $\mathcal{L}$ of the architecture is defined as:
\begin{equation}
    \mathcal{L} = \lambda_1\mathcal{L}_{loc_{1}} + \lambda_2\mathcal{L}_{loc_{2}} + \lambda_3\mathcal{L}_{class}
\end{equation}
,where $\lambda_i$ are the weights of the partial loss functions. To focus the training on the localization and to make them on the similar magnitude, we set $\lambda_1 = 50$, $\lambda_2 = 100$ and $\lambda_3 = 1$.
Our hyperparameter optimization of the Stochastic Gradient Descent (SGD) optimizer determined the learning rate as $5\cdot10^{-4}$, the momentum as 0.9, and a weight decay of $10^{-5}$ as the best configuration for our training.  
A further difference is that our architecture includes dropout~\cite{dropout} with an extinction probability of $0.2$ for the input and the first convolution of the backbone and dropout with a probability of $0.5$ for each additional convolution layer in the backbone and for the first three convolution layers of the \emph{RepPoints} architecture. 

As a post-processing, to keep only the most confident predictions, we applied Non-Maximum Suppression (NMS) with a threshold of 0.4 IoU to keep boxes.

\subsection{Illumination Augmentation}\label{sec:illuminationAugmentation}
\begin{figure}[t]
\begin{center}
   \includegraphics[width=0.8\linewidth]{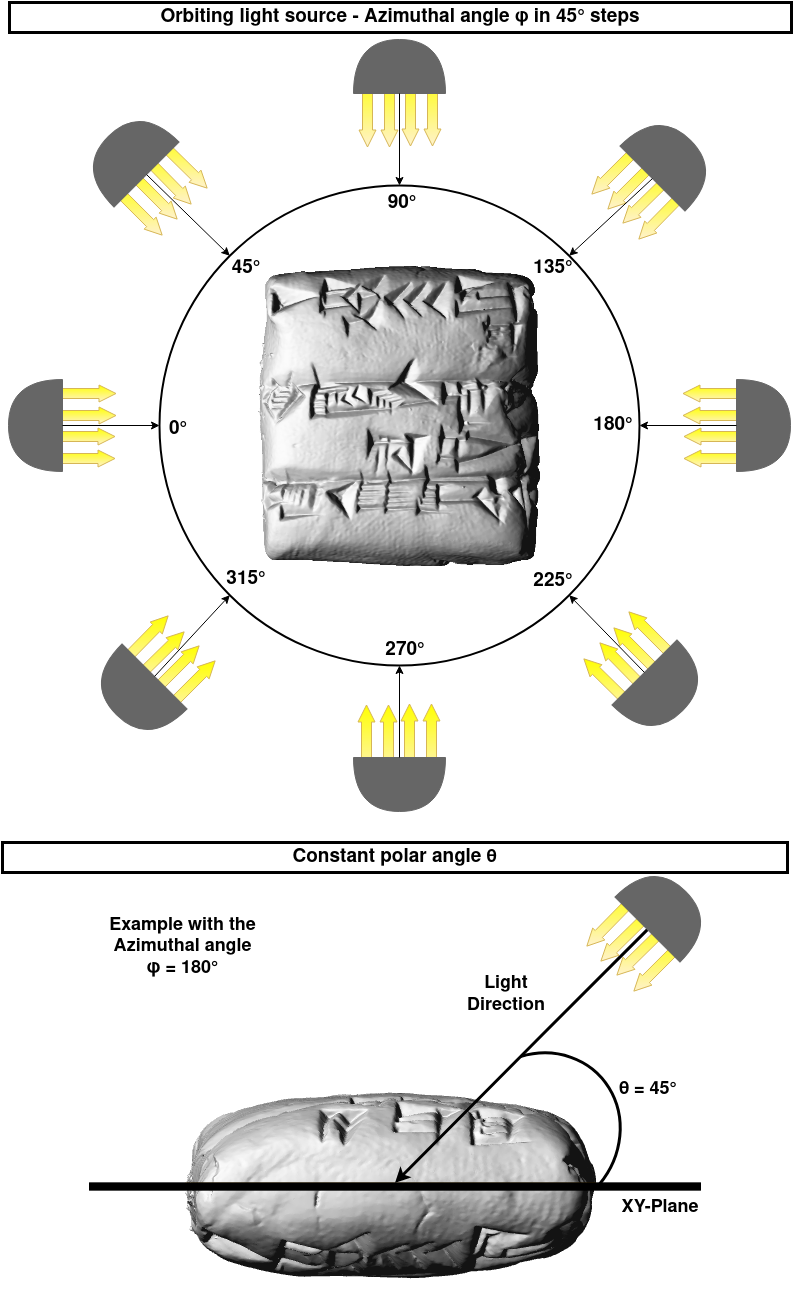}
\end{center}
   \caption{Orbiting Light Source to render the clay tablets under different illumination conditions. The upper part of the diagram illustrates the front view of the tablet, around which the direction light rotates 360°. The side view shown below is only for a better 3D orientation and visualizes the constant polar angle. The tablet shown is \href{https://doi.org/10.11588/heidicon/1113625}{HS 1194}.}
\label{fig:illuminationMeth}
\label{fig:onecol}
\end{figure}

Since the shape of the Cuneiform signs is three-dimensional, they appear differently depending on the type of illumination. Additional to the direct impact of the brightness, the sign appearance depends strongly on the angle of incidence. The wedge shadows vary due to their curvature if the light is not orthogonal to the tablet front. Similar to \cite{Rest22}, we used this characteristic of Cuneiform to render our available tablets with different illumination and have augmented our data set with a huge set of virtual light renderings. Our approach to illumination augmentation~(\emph{IA}) is shown in \ref{fig:illuminationMeth}. We used the open-source \emph{GigaMesh Software Framework} to render the meshes of the clay tablets under a virtual light (Phong). Since only annotations of the back and front are available, these sides are rendered with an orbiting light source, with the azimuth angle $\phi$ varying in 45° increments from 0° to 360°. To avoid data overload, we set the polar angle $\theta$ constant to 45° (according to \cite{Rest22}).

\section{Data}\label{sec:data}

The methods described above are applied to the \emph{Frau Professor Hilprecht Collection
of Babylonian antiquities} at the University of Jena, which is published as 3D data in combination with high-resolution renderings by the \emph{Heidelberg Cuneiform Benchmark Dataset})~(HeiCuBeDa)~\cite{mara2019heicubeda}. The dataset consists of three different types of 3D renderings: \texttt{VirtualLight}~(VL), \texttt{MSII filter}~(MSII), and a mixture of both~(mixed).
The VL is a Phong rendering and the default technique of the \emph{GigaMesh Software Framework}\footnote{https://gigamesh.eu} and can be seen in \cref{fig:resultVL}.
Increasing the contrast between the impressed wedges and the surface based on the curvature is done by the MSII filter. This type of rendering was used to create \cref{fig:repPoints1,fig:repPoints2}.
The subset of tablets used, with the exception of one tablet from the Old Babylonian period (c. 1900-1600 BC), is dated between 2500 and 2000 BC. Except for one tablet in Akkadian, all tablets are written in Sumerian.

The annotations of these renderings were made using the Cuneiform Annotator application \emph{Cuneur}~\cite{cuneur22} and published as the \emph{Mainz Cuneiform Benchmark Dataset}~(MaiCuBeDa)\footnote{MaiCuBeDa: \url{https://doi.org/10.11588/data/QSNIQ2}}. 
Due to our focus on sign detection in this work, we only applied the sign annotations of the dataset, ignoring the sign classes (the Unicode code point of the cuneiform sign), which were deemed irrelevant for this task. These annotations are based on the transliterations available at the Cuneiform Digital Library Initiative~(CDLI)~\cite{englund2016cuneiform} of the Hilprecht collection. 
Each annotated sign of MaiCuBeDa refers to a sign in the transliteration. Unfortunately, if a sign in the image could not be classified and assigned, it led to missing sign annotations within the dataset. In particular, the side signs of the tablet are often not annotated, but there are also cases of missing annotations in the center of a tablet since transliterations may be incomplete or signs could not be clearly assigned by the annotating person. As a result, we have used a challenging dataset that contains predominantly incompletely annotated images. 

\begin{table}[h]
\begin{center}
\begin{tabular}{l|c}
Number of tablets& 490 \\
\hline
Number of segments& 873 \\
\hline
Number of $512\times512$ pixels patches & 10311 \\
\hline
Number of sign annotations& 21228 \\
\end{tabular}
\caption{Our available annotated data}
\label{tab:dataset}
\end{center}
\end{table}

As part of this work, several extensions and preprocessing of the dataset were performed.
First, due to the different shapes of the clay tablets, the corresponding images have different resolutions. To standardize and to avoid loss of information when rescaling to the input resolution described in \cref{sec:methSignDet}, we crop the original images into patches of $512\times512$ pixels with an overlap of 256 pixels. The \Cref{tab:dataset} provides an overview of the resulting patches, the original number of segment images, and the available annotations. 
%To get an overview of the resulting patches, the original number of segment images and the available annotations, see \cref{tab:dataset}. 
In addition to the originally provided renderings, applying the \emph{IA} described in \Cref{sec:illuminationAugmentation} extended the VL image set by 7344 additional images of whole segments and thus 82488 overlapping patches to an image set of 92799 VL rendering patches. 
We also added the corresponding photographs available at the CDLI to the dataset. Since these photographs do not directly match the renderings, we mapped the images using the \emph{Cuneur Transformer}\footnote{Cuneur-Transformer:\url{https://gitlab.com/fcgl/cuneur-transformer}} tool so that the annotations created for the renderings could also be used for the photographs. Since all types of renderings are grayscale, the photographs are converted to grayscale and performed a normalization. This is done to standardize the input to the CNN.
\section{Results}\label{sec:results}
In this section, we introduce our evaluation method in \Cref{sec:evaluation}, which is used in \cref{sec:resSignDet} to describe our results of sign localization. These results are discussed in \Cref{sec:discussion}.  

\subsection{Evaluation}\label{sec:evaluation}
For each of our experiments, we divide the same training, validation, and test dataset with a ratio of $2:1:1$. To avoid overlap between the datasets at the level of sign clippings and cropped squares, we initially split the dataset at the segment level and then we crop the images based on this split. All evaluations are carried out on the separate test set by the models that have performed best on the validation set that is evaluated at the end of each epoch.  

To evaluate our sign detector, we use the Average Precision (AP), which is a common evaluation metric for object detection~\cite{2019LiuSurvey}. Varying the confidence threshold for deciding whether a prediction is a sign or background results in different precision and recall values per threshold, thus yielding a precision/recall curve. As defined in \cite{Everingham2010ThePV}, we use the interpolated precision/recall curve with 11 recall levels between 0 and 1 to calculate the AP. Furthermore, we vary the threshold of the IoU between the predicted bounding box and the GT bounding box $\theta_{IoU}$ which determines whether a detection is classified as true positive or as false positive, to evaluate the localization accuracy of the bounding boxes. In the following, AP@$\theta_{IoU}$ notes the thresholds in percent used during the evaluation, \eg, AP@50 represents an evaluation where 50\% IoU overlap is required for a bounding box to be considered as true positive.

\subsection{Sign Detection}\label{sec:resSignDet}

\begin{table}[h]
\begin{center}
\begin{tabular}{p{1.6cm}|p{1.2cm}||p{1.0cm}|p{1.0cm}|p{1.0cm}}
\textbf{Train set}& \textbf{Test set}& \textbf{AP@50} & \textbf{AP@75} & \textbf{AP@90} \\
\hline
\multirow{4}{1.6cm}{VirtualLight renderings} & Photos & 0.327 & 0.182 & 0.182\\
\cline{2-5}
& VL & \textbf{0.570} & \textbf{0.362} & 0.182\\
\cline{2-5}
& MSII & 0.503 & 0.268 & 0.182\\
\cline{2-5}
& Mixed & \textit{0.555} & 0.288 & \textbf{0.273}\\
\hline
\hline
\multirow{4}{1.6cm}{MSII renderings} & Photos & 0.288 & 0.182 & 0.145\\
\cline{2-5}
& VL & 0.375 & 0.212 & 0.182\\
\cline{2-5}
& MSII & \textbf{0.602} & \textbf{0.437} & \textbf{0.273}\\
\cline{2-5}
& Mixed & \textit{0.575} & 0.357 & \textbf{0.273}\\
\hline
\hline
\multirow{4}{1.6cm}{Mixed renderings} & Photos & 0.244 & 0.182 & 0.071\\
\cline{2-5}
& VL & 0.443 & 0.222 & 0.133\\
\cline{2-5}
& MSII & \textit{0.591} & \textit{0.400} & \textbf{0.182}\\
\cline{2-5}
& Mixed & \textbf{0.609} & \textbf{0.420} & \textbf{0.182}\\
\hline
\hline
\multirow{4}{1.6cm}{Photos} & Photos & 0.456 & \textit{0.222} & \textbf{0.182}\\
\cline{2-5}
& VL & \textit{0.517} & \textit{0.269} & \textbf{0.182}\\
\cline{2-5}
& MSII & \textit{0.514} & \textit{0.267} & \textbf{0.182}\\
\cline{2-5}
& Mixed & \textbf{0.523} & \textbf{0.287} & \textbf{0.182}\\
\hline
\hline
\multirow{4}{1.6cm}{VirtualLight renderings with \emph{IA} } & Photos & 0.417 & 0.213 & 0.144\\
\cline{2-5}
& VL & \textbf{0.603} & \textbf{0.452} & \textbf{0.364}*\\
\cline{2-5}
& MSII & \textit{0.585} & 0.363 & 0.170\\
\cline{2-5}
& Mixed & \textit{0.583} & 0.365 & 0.152\\
\hline
\hline
\multirow{4}{1.6cm}{Complete} & Photos & 0.508 & 0.229 & 0.132\\
\cline{2-5}
& VL & 0.598 & 0.368 & \textbf{0.182}\\
\cline{2-5}
& MSII & \textit{0.632} & \textbf{0.394} & 0.160\\
\cline{2-5}
& Mixed & \textbf{0.633} & \textbf{0.394} & \textit{0.177}\\
\hline
\hline
\multirow{5}{1.6cm}{Complete with \emph{IA}} & Photos & 0.569 & 0.214 & 0.170\\
\cline{2-5}
& VL & \textit{0.626} & \textbf{0.545}* & \textbf{0.182}\\
\cline{2-5}
& MSII & 0.591 & 0.400 & \textbf{0.182}\\
\cline{2-5}
& Mixed & \textbf{0.636}* & 0.431 & \textbf{0.182}\\

\end{tabular}
\caption{Results of the Sign Detector on cropped $512\times512$ pixel-sized patches compared with different train and test sets. Complete means the combination of mixed renderings, MSII filter renderings, VL renderings, and photographs.  The best results of each training data set are highlighted in bold and those close to the best are highlighted in italics. The best results per evaluation method are marked with an asterisk *.}
\label{tab:resultsDifDatasets}
\end{center}
\end{table}

The results take into account several aspects that affect the performance of the sign detector: the role of various training data, the applicability to different image types, and the impact of \emph{IA}.   
The \cref{tab:resultsDifDatasets} provides an overview of all results for models trained with different training bases and evaluated on various test image sets. Having the same hyperparameters, they only differ in their input images, after which epoch the best model has been evaluated on the validation set. The models trained on a combination of various image types or \emph{IA} renderings achieved the best-performing model in fewer epochs compared to training on a single-source dataset.  

The results of the pure datasets have shown that the best evaluation results are obtained when the applied image type for training is the same as for testing. One exception is the training on photographs, which results in a model that performs better on all types of renderings; however, these results are close to each other. Conversely, the models trained on renderings achieve a lower average precision on photographs. Even the best result with VL renderings is over 0.1 lower than a model trained directly on photographs. 
%Another notable result of applying pure datasets is that mixed renderings achieve much better results on VL than MSII renderings.

Applying a combination of the photographs and all types of renderings slightly increased the performance of each image type compared to the best models trained on pure datasets. Thus, the model is generally applicable to each type of image presented in this work because it is able to detect the signs as well as the best model trained on the respective image set.

The use of \emph{IA}, as described in \cref{sec:illuminationAugmentation}, only as different VL renderings, has shown a slight improvement compared to the evaluation on VL renderings for AP@50 and AP@75, but it has increased the AP@90 by about 20\%. 
%Furthermore, there is an improvement on the photographs
Although this strength in AP@90 was not observed with the training on the combination of all types of images and the additional VL renderings by \emph{IA}, this model achieved the highest AP@75 on VL renderings, the highest AP@50 on photographs and the highest AP@50 on mixed renderings. However, the result on the mixed renderings is close to the combination without \emph{IA}. 
%To place the results into a different metric, this model achieved a $f_1-Score$ with an 50\% IoU threshold of \textbf{??} on mixed renderings and \textbf{??} on photographs.

\begin{figure}[t]
\begin{center}
   \includegraphics[trim={0 80 0 70},clip,width=0.8\linewidth]{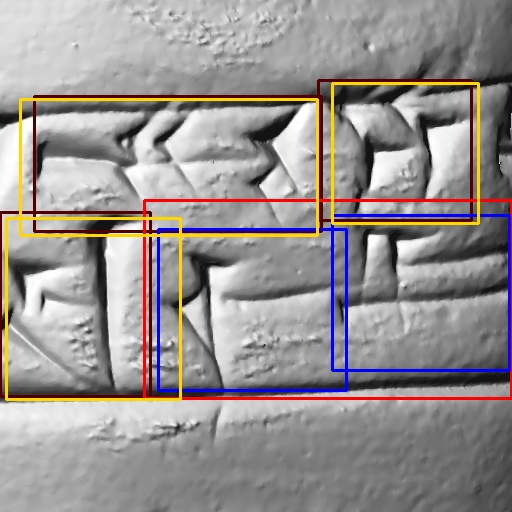}
\end{center}
   \caption{Sign detection result on a VL rendering patch using the model trained on VL renderings with \emph{IA}. This example shows the GT bounding boxes (black), true positives (yellow), false negatives (red), and false positives (blue) of the evaluated patch.}
\label{fig:resultVL}
\end{figure}

In general, the gap between the AP@50 and the AP@75 is tiny when the sign detector is applied to the renderings, but the AP@90 is even smaller. However, the strict 90\% IoU overlap restriction is not representative because of the dependence on the GT bounding box. Some bounding boxes are often not very close to the boundary of the sign, so some very close detections will not be considered true positives.
Although not every GT bounding box exactly surrounds the signs, the difference between AP@50, AP@75, and AP@90 indicates how accurately the bounding boxes are predicted. Consequently, the high AP@75 values of the evaluation on the renderings suggest an accurate prediction of the bounding box. Our visual investigations of the results came to the same conclusion that the predicted bounding boxes are close to the signs when we use renderings as inputs. This can be seen for one example on VL renderings in \cref{fig:resultVL}. In this figure, a weakness of the sign detector can also be seen: the detector tends to predict signs as small units and, consequently, to split larger compound signs. This results in the two false positives (blue) of the small units and one false negative (red) of the wide ground truth sign.   
In addition, the model detects sometimes false positives on seals and the broken surface of tablets, and it rarely mistakes the written tablet identification number for a sign in the photograph.

\begin{figure}[ht]
\begin{center}
   \includegraphics[width=0.8\linewidth]{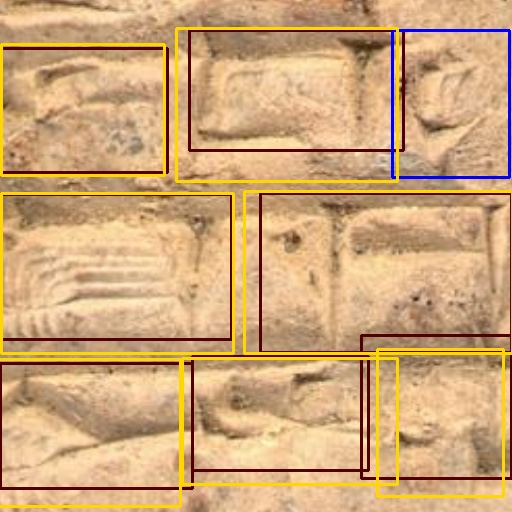}
\end{center}
   \caption{Sign Detection result on a photograph patch using the model trained on the combination of the datasets with \emph{IA}. This example shows the GT bounding boxes (black), true positives (yellow) and one false positive (blue) of the evaluated patch.}
\label{fig:resultPhoto}
\end{figure}

As mentioned before in \Cref{sec:data}, the tablets are incompletely annotated, so there are a lot of false positives, which are, in reality, signs. Consequently, the actual results are better than the numbers. The \Cref{fig:resultPhoto} shows one of these false positives. 

\subsection{Discussion of results} \label{sec:discussion}

As described in \Cref{sec:resSignDet}, the sign detection achieved sufficient results, especially considering our dataset's missing ground truth annotation.
To rank our results, we compare them with a state-of-the-art cuneiform sign detector, which performs the same task as ours on Elamite tablets \cite{23deepScribe}, in \cref{tab:compareDeepScribe}. 
%\cite{23deepScribe} are blessed with a huge dataset compared to our. In terms of numbers, our dataset contains only about 18\% as many sign annotations as theirs.
Due to the large difference in size of the dataset, which contains only about 18\% as many sign annotations as \cite{23deepScribe}, all results must be put into perspective.
%115921 williams signs 21228 our signs
Our approach achieved a lower AP@50 for detection in photographs but a similar AP@75. However, due to the absence of annotations, the actual performance is better. Furthermore, our annotations were not created for these photographs, so due to the transformation to process the mapping by the \emph{Cuneur Transformer}, there may be small deviations in the annotated location, and some lateral signs are generally not visible in the photographs. Since we do not have a dataset with the annotations specifically created for the photographs available, we are unable to measure the impact of the transformation error and its associated effects on the sign detector, but according to our visual evaluation of the \emph{Cuneur Transformer}, there are only a few examples with incompletely matching bounding boxes for the signs.

\begin{table}
\begin{center}
\begin{tabular}{l|c|c}
Approach & AP@50 & AP@75 \\
\hline\hline
DeepScribe\cite{23deepScribe}&  0.77 & 0.21 \\
\hline
%\cite{23deepScribe} 10\% of the dataset& $\approx$ 0.65 & not given   \\
Ours for mixed renderings & 0.64 & 0.43\\
Ours for VirtualLight renderings & 0.63 & 0.55\\
Ours for photographs & 0.57 & 0.21\\
\end{tabular}
\end{center}
\caption{Sign detection result compared with \cite{23deepScribe}. Our model was trained with all types of renderings, photographs, and augmented VL renderings through \emph{IA}. The model was evaluated for different input images.}
\label{tab:compareDeepScribe}
\end{table}

According to our results, the utilization of renderings to train a Cuneiform machine learning model seems to be a suitable approach.  
Hence, despite the smaller amount of data with incomplete sign annotations, our AP@50 is close to the result in \cite{23deepScribe}. In addition, our AP@75 on renderings exceeds \cite{23deepScribe} results by as much as 0.3 due to our accurately placed bounding boxes. For a pipeline approach like the ones described in \Cref{sec:pipeline} or \cite{23deepScribe}, where the signs are cut out to process them in a subsequent step, it could lead to the loss of essential information by locating only partial signs. Therefore, our approach could have an advantage over such a pipeline. Presumably, our approach could achieve the same results if a fully annotated dataset with more signs were available for training and testing.

As described in \Cref{sec:resSignDet}, dealing with compound-wide signs has proven to be a challenge. Due to the wide range of sign interpretations, it is even difficult for a human to determine the sign boundaries. Consequently, there is possibly no consistent dataset available.

 Considering various methods of rendering a mesh, such as with the MSII preprocessing, has shown that transferring 3D information to a 2D image can improve the detector performance. Using the MSII filter on the meshes results in a higher contrast between the wedges and the clay surface based on the curvature. 3D renderings after MSII filtering are much more legible than photographs, which is consistent with human perception of cuneiform tablets. In addition to preprocessing, a mesh of a clay tablet offers the possibility of \emph{IA}, as described in \Cref{sec:illuminationAugmentation}. Applying this method to augment the dataset has shown an improvement in the sign detector for VL renderings. Specifically, an increase of about 0.1 of AP@75 and 0.2 of AP@90, indicates that the accuracy of the bounding box has improved.

Despite the better results with the renderings, the best results were obtained in combination with the photographs. This could be explained by the different information provided by renderings and photographs, which helps the model to generalize. Consequently, it seems necessary to use both media to get good machine learning OCR results.

\section{Conclusion and Outlook}
We have investigated a Cuneiform sign detector based on \emph{RepPoints} to locate signs and cut them out for subsequent steps, as in the pipeline approach of \cite{23deepScribe} or \cite{our23}, from the viewpoint of the bounding box localization accuracy, the dataset impact and the improvement by \emph{IA}. Achieving a high AP@75 on renderings suggests that the detector's proposed bounding boxes completely encircle the signs, which is necessary to cut them out for a previous step in a pipeline. In further research, it would be interesting to apply this sign detection with a rendering training dataset in the pipeline of \emph{DeepScribe}~\cite{23deepScribe} to see if the classification result can be improved by more accurate bounding boxes and by the renderings themselves.
At the moment, the signs are only localized in the form of a bounding box; however, the representation points of the \emph{RepPoints} might be able to represent them differently.
Varying the backbone, pre-training the architecture with a different dataset, or increasing the number of ResNet~\cite{ResNet} layers is another way to study the architecture. 

Furthermore, our research has shown that using 3D scans offers a wide range of possibilities. First, the meshes provide the ability to apply \emph{IA}, which has shown improved results and can be scaled by angle variation in the future. Second, our results suggest that a preprocessing of the meshes, in our case MSII, also increases the performance of the sign detector. Further research could also evaluate other algorithms as mesh preprocessing, \eg \emph{Ambient Occlusion}. However, it should be noted that 3D model datasets are rarely available and are time-consuming to create due to the 3D scanning process.

To compare the performance between the originally unannotated photographs and the different types of renderings, we mapped them onto the renderings to make them accessible for the available annotations. 
Our results have shown: All types of renderings can produce better results than photographs.
Although the mixed renderings are the most suitable input to localize signs, the best results have been achieved by the training with a combination of all datasets, including photographs.
As our research has shown, the combination of 3D scans with photographs provides a great opportunity to create and improve  machine learning models of cuneiform OCR.

\subsection*{Future Work}
\label{sec:outlook}
%Cuneiform sign location and classification is a crucial step to work toward the goal of automatically transcribing cuneiform tablet images in the form of transliterations or translations or even their application in augmented reality settings such as Google Lens, which could bring the automated analysis of cuneiform tablet contents to a new level. 
The localization and classification of cuneiform signs is a crucial step towards the goal of automatic transcription of cuneiform tablet images in the form of transliterations or translations, or even their application in augmented reality environments such as Google Lens, which could take automated analysis of Cuneiform tablet content to a new level.
While this work has given insights into which types of media and their combinations can improve the classification and location tasks, future experiments could tackle the combination of sign classification approaches with transliteration assignments or automated translation approaches. Also, one could think of repeating the experiments of this work with cuneiform tablet renderings of cuneiform tablets of epochs that were not considered by the MaiCuBeDa dataset or previous datasets to discover epoch, language, or writing style specific challenges. 
Finally, the location and classification of not only cuneiform signs but also of paleographic sign variants, potentially varying even in the same spatio-temporal settings, will be a research challenge of great importance for creating and linking to accurate cuneiform paleography databases, such as the emerging PaleOrdia \footnote{\url{https://situx.github.io/paleordia/script/?q=Q401&qLabel=cuneiform}} based on Wikidata.
%In the future, it could be beneficial to base machine learning datasets for Cuneiform OCR tasks on 3D scans and photographs. 

\section*{Acknowledgements}
We thank Florian Linsel and Jan Philipp Bullenkamp for their editorial assistance.
This work was partially funded by DFG Project "Die digitale Edition der Keilschrifttexte aus Haft Tappeh" under project reference number 424957759 \footnote{\url{https://gepris.dfg.de/gepris/projekt/424957759}}

{\small
\bibliographystyle{ieee_fullname}
\bibliography{egbib}
}

\end{document}